# Chinese Traditional Poetry Generating System Based on Deep Learning

Chenlei Bao，Lican Huang

College of Informatics，Zhejiang Sci-Tech University, Hangzhou 310018,China

598562464@qq.com , Licanhuang@zstu.edu.cn

**Abstract**   Chinese traditional poetry is an important intangible cultural heritage of China and an artistic carrier of thought, culture, spirit and emotion. However, due to the strict rules of ancient poetry, it is very difficult to write poetry by machine. This paper proposes an automatic generation method of Chinese traditional poetry based on deep learning technology, which extracts keywords from each poem and matches them with the previous text to make the poem conform to the theme, and when a user inputs a paragraph of text, the machine obtains the theme and generates poem sentence by sentence. Using the classic word2vec model as the preprocessing model, the Chinese characters which are not understood by the computer are transformed into matrix for processing. Bi-directional Long Short-Term Memory is used as the neural network model to generate Chinese characters one by one and make the meaning of Chinese characters as accurate as possible. At the same time, TF-IDF and TextRank are used to extract keywords. Using the attention mechanism based encoding-decoding model, we can solve practical problems by transforming the model, and strengthen the important information of long-distance information, so as to grasp the key points without losing important information. In the aspect of emotion judgment, Long Short-Term Memory network is used. The final result shows that it can get good poetry outputs according to the user input text.

## 1 INTRODUCTION

Chinese traditional poetry is an important intangible cultural heritage in China. It contains many life philosophies and historical stories from ancient times to now. It is an artistic carrier of thought, culture, spirit and emotion. However, because ancient poetry has strict rules and restrictions, it is very difficult to generate it by machines. In recent years, great breakthroughs have been made in the related technologies of deep learning, which has been widely and successfully applied to various aspects such as speech recognition, image recognition, natural language processing and so on. The automatic generation of poetry has attracted more attention[1,4]. The machine should not only learn the structural characteristics, rules of poetry, but also learn the various meanings of Chinese characters and the author's emotion contained in the characters. This paper mainly studies the application of deep learning technology to the automatic generation of Chinese traditional poetry.

The automatic generation of poetry in western countries is earlier than that in China. However, the early machines could only use simple programs to combine a pair of words or copy them mechanically, and did not have the internal meaning of ancient poetry and songs. The main technical research work of mechanical poetry generation began in the 1970s. The following introduces several traditional poetry generation methods.

Word Salada method is the earliest method of poetry generation. This method takes words as a kind of material and mixing words randomly. Because this combination does not consider the meaning of words and the requirements of poetry, the poetic effect is poor.

Poetry generation method based on Template and pattern is similar to the cloze filling in sentences. By setting some template formats in advance, digging out the subject, predicate, etc., using the original text as the poetry template, and then filling in the blanks with words in line with this form, a new poem can be generated. Using this method to generate words can avoid the problem that some words do not consider the nature of words and poetry, but it is very rigid because it is a way to fill in the blank. Later, a pattern based generation method appeared. This mode is to generate poetry under the set mode, and effectively generate poetry by limiting the characteristics of each position word and the tonal rhyme and oblique tones of poetry, which can meet the requirements of grammar and rhythm to a certain extent.

Changle Zhou, et.al[2] proposed a method to generate poetry based on genetic algorithm. It regards the poetry generation process as a search, starts with a random line, uses a predetermined evaluation method to evaluate, and then iterates, and finally can get the desired poem. This method can better generate a poem line, but it has no good effect on the whole poem.

Rui Yan, et.al [20] were inspired by automatic summarization technology and applied it to poetry generation. Through the words entered by the user, the system searches and selects similar words in the corpus, sorts and selects the keywords according to the conditions of semantics, tonal rhyme, etc., and finally modifies and replaces them through the optimization of restrictive conditions to obtain poetry.

Jing He, et.al [3] regard poetry generation as a machine translation problem, take the previous sentence as the prediction input of the next sentence, add constraints, predict the next sentence, and cycle the process until a complete poem is generated.

## 2  Related Work

### 2.1  Word vector model

Text is a kind of unstructured data information, which cannot be directly calculated and used. The function of text representation is to transform unstructured information into structured information, so that we can effectively perform calculations on these text information.

Word Embedding can be understood as a kind of mapping, in which words are calculated, transformed, and projected into another space that can be represented by numbers and can be understood by computers.

Word2vec[5,6,7] is one of the methods of Word Embedding in which through the function f(x)->y, the text is transformed into a vector. Use the weight matrix that is gradually improved during the training process to transform the words into a one-dimensional matrix, which is the word vector.

Word2vec has two main methods, CBOW (Continuous Bag-of-Words Model) and Skip-Gram (Continuous Skip-gram Model)[8]. Among them, CBOW predicts the current value through the context, and Skip-Gram: predicts the context through the current value.

### 2.2  Keyword extracting

The key word extraction methods for text mainly include the following three, which are called supervised, semi-supervised and unsupervised.

1. Supervised keyword extraction: The data set has already marked keywords in advance.

2. Semi-supervised keyword extraction: provide part of the labeled data for training, and then add the words obtained in the new data to the data set after confirmation, and proceed in a loop.

3. Unsupervised keyword extraction: directly extract the given text[9].

This article mainly uses two unsupervised keyword extraction methods, TF-IDF[11] and TextRank[10].

#### 2.2.1 TF-IDF

The main idea of TF-IDF is: if a word appears frequently in an article and less frequently in other articles, then the word has a good classification ability for the article. Among them, TF means word frequency, that is, the number of occurrences divided by the total number of words in the article; and IDF means the degree of discrimination. The degree of discrimination means that the lower the probability of a word, the better the effect it can achieve, and the higher the degree of discrimination. The keyword is with a large TF-IDF value.

Suppose a word w appears a times in the article, and there are b words in this article, then

$TF_w = \frac{a}{b}$ (1)

IDF represents the frequency of occurrence. Assuming that the entire corpus has a total of C texts, of which B contains the word A, then

$IDF_A = \log_2 \frac{C}{B}$ (2)

Therefore, the TF-IDF value of word w can be obtained:

$TFIDF_w = TF_w * IDF_w$ (3)

### 2.2.2 TextRank

The TextRank algorithm is a graph-based ranking algorithm for text. By dividing the text into several units and building a graph model, using the voting mechanism to sort the important components of the text, and using the information of the article itself, keyword extraction can be achieved.

TextRank can be expressed as a directed weighted graph G=(V, E), composed of a point set V and an edge set E, where E is a subset of V*V. In the graph, the weight of the edge between any two points Vi and Vj is Wji. For a given point Vi, In(Vi) is the set of points pointing to Vi, and Out(Vi) is the set of points Vi pointed. Vi's score is calculated as follows:

$WS(V_i) = (1-d) + d * \sum_{V_j \in In(V_i)} \frac{w_{ji}}{\sum_{V_k \in Out(V_j)} w_{jk}} WS(V_j)$ (4)

### 2.3 Network Model BiLSTM

Recurrent neural network RNN[12,13,14,15,16,17,18] means that the network will repeat the same action over time. RNN has a memory function and has a very good effect on time series analysis. The characteristics of RNN: the current state at time t will be affected by the state at time t-1.

Many scenarios require more context. Under normal circumstances, there will be intervals between the meanings of related sentences. As the interval increases, the prediction accuracy of RNN in this aspect begins to decline. At this time, the Long Short-Term Memory network(LSTM) can be used. LSTM belongs to RNN and has the characteristic of maintaining long-distance connections.

Although the LSTM model can handle the sentence model better, there is still a problem, that is, it cannot handle the information from the back to the front. At the same time, when dealing with more detailed content, more attention should be paid to the interaction of emotion words and degree words.

BiLSTM is a combination of forward and reverse LSTM. Combine the two sets of vectors obtained in the forward and reverse directions to obtain the final vector.

For example, the vectors $h_{L0}$, $h_{L1}$ and $h_{R0}$, $h_{R1}$ are obtained in the forward and reverse directions, respectively. Combine the two sets of vectors to get the vector combination [$h_{L0}$, $h_{R0}$], [$h_{L1}$, $h_{R0}$], that is, $h_0$, $h_1$.

### 2.4 Encoder-decoder model

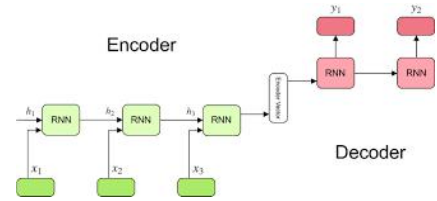

Fig.1 Encoder-decoder model

In the selected network model, the hidden layer state $h_t$ at the current moment is not only determined by the input $x_t$ at the current moment, but also by the hidden layer state $h_{t-1}$ at the previous moment.

$h_t = f(h_{t-1}, x_t)$ (5)

In the encoder stage, we sum the states of the hidden layers to generate the final sum semantic encoding vector C.

$C = q(h_1, h_2, h_3, …, h_{Tx})$ (6)

The last hidden layer state is taken as the semantic encoding vector C.

In the decoding stage, the next output word y is predicted according to the given semantic vector C and the generated output sequence $y_1, y_2, …, y_{t-1}$.

$y_t = \mathrm{argmax} P(y_t) = \prod_{t=1}^{T} p(y_t | y_1, y_2, …, y_{t-1}, C)$ (7)

It can be abbreviated as:

$y_t = g(y_1, y_2, …, y_{t-1}, C)$ (8)

## 2.5 Emotion analysis

Emotion analysis [19] is mainly divided into the following processes:

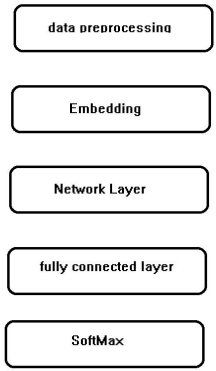

Fig.2 Emotion analysis process

Data preprocessing includes removing garbled characters, removing stop words, and segmentation words. Because the length of each sentence is different, and most ancient poems are five-character or seven-character, you can choose the longest of the current bach_size as the benchmark, and fill the length with 0 to the length of the current bach_size to facilitate the follow-up of the data deal with.

The Embedding layer converts one-dimensional vectors of words into high-dimensional vectors in order to obtain more semantic information. This article uses 300 dimensions as the high-dimensional dimensions.

The network layer is initialized first, and then the input vector is superimposed with the initial hidden state, the weight is updated through the network, and the hidden state $h$ is obtained. The previously extracted features are synthesized through a fully connected layer, the information after the integration result is output through a certain linear operation, and the processed data is used for classification.

The Softmax output layer classifies n categories. The input of the fully connected layer is passed through the Softmax function to obtain n probability values, the largest of which is the result of the classification.

## 3 Chinese Traditional Poetry Authoring System Implementation

### 3.1 poetry data source and data preprocessing

The poetry data comes from the GitHub project Chinese Poetry[22]: the most complete database of Chinese Poetry Classics, including hundreds of thousands of ancient poems and songs of the Tang and Song dynasties in JSON format.

Because ancient poetry is mainly divided into five character and seven character poetry , ancient poetry is first separated into two parts according to five character and seven character .

Read the five character and seven character data respectively, obtain the poetry content, obtain the keyword of each poem, and store it in a new file according to the format of keyword + current poem and existing previous text. For example, the poem "床前明月光，疑是地上霜(Moonlight in front of bed, suspected to be frost on the ground)" is stored as "月光|床前明月光(Moonlight | moonlight in front of bed)" and "霜|床前明月光，疑是地上霜(frost | moonlight in front of bed, suspected to be frost on the ground)".

Set four special character marks, namely, space, unknown, start and end, to facilitate the determination of the beginning and end of poetry, special characters and other unqualified contents during prediction and training. Traverse the data and build a sequence index .

Select the maximum sequence length in each batch as the benchmark of the current batch, and fill the remaining sentences less than the sequence length with spaces until the length is reached.

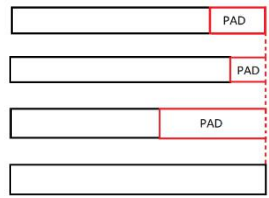

Fig.3 Completing sequence with PAD

## 3.2 Chinese word vector

Chinese Wikipedia is used as the training set, and Skip-Gram method (word2vec) is used for training. The method based on context co-occurrence and PCA dimension reduction is used for dependency vectors. The number dimension of the trained word vector is 20029 and the dimension is 300.

In poetry training, the whole pre-trained file is read. If there is a word that is not in the trained word vector file, the word is randomly assigned a value suitable for the dimension and added to the word vector. In subsequent training, the value of the word is adjusted to make it conform to the position of word meaning in space as much as possible.

## 3.3 Construction Network Model

Read the pre trained data, build id2word and word2id sequences, and build embedding.

For the encoder side, use the built embedding to obtain the keyword and the previous content data in the current batch. The bidirectional long-term and short-term memory network (BiLSTM) is used as the network layer to create neurons with two keywords and previous content for forward and backward respectively. Then build the network of keywords and the previous content respectively. In the neural networks of the keyword and the previous content, the output $h_L$ and $h_R$ obtained in the forward and backward directions are spliced and combined into the final h, and the dimension is specified as 1. Finally, the key words and the output of the previous content are spliced to form a whole.

For the decoder side, the same as the encoder side, use the built embedding and get the contents in the batch. The functions used for training at the decoder are TrainingHelper and GreedyEmbeddingHelper respectively. TrainingHelper does not take the output of t-1 stage as the input of t-stage, but directly inputs the real value to RNN, so that each input is the real value of text. For example, "床前明月光(bright moon in front of bed)", the input of TrainingHelper is ['床', '前', '明', '月', '光' ('bed', 'front', 'bright', 'moon', 'light')]. GreedyEmbeddingHelper is generally used for decoder decoding in the prediction stage. It does not use the real value of the text, but embeds the value obtained from each prediction as the next input value. The difference between GreedyEmbeddingHelper and TrainingHelper is that it embeds the output under t-1 and then inputs it to RNN.

## 3.4 Model Training

a) Select loss function sequence_loss, the calculation steps are: 1) Softmax evaluation, 2) Cross entropy selection, and 3) Calculate Average.

b) Use the Adam optimizer.

c) Calculate the gradient.

d) The calculated gradient is used to update the corresponding variable.

## 3.5 Prediction Implementation

The poetry generation method is: the user inputs a paragraph of text the user input is extracted as the key words to form a poetry outline. Each verse to be generated is matched with a keyword to ensure that the poetry content does not drift. If the keywords extracted from the text segment entered by the user are not enough to generate the required number of poem lines, the keywords will be expanded by looking for synonyms.

During prediction, the keywords extracted from user input are sequentially input into the computing diagram, and the preceding text of the first keyword is empty. Then add the generated poem line to the previous list in turn, and input them into the computing diagram together with the keywords in order. For example, there are keywords "月光(Moonlight)" and "霜(frost)". First input "月光(Moonlight)" and empty preamble to obtain the first line "床前明月光(Moonlight in front of bed)"; Then input the keyword "霜(frost)" and the preceding text "床前明月光(Moonlight in front of bed)" to obtain the second line "疑是地上霜(suspected to be frost on the ground)".

## 3.6 Evaluation of Model

This paper uses the perplexity degree to judge the quality of the model. The basic idea of perplexity degree is: the model which assigns a higher probability value to the sentences in the test set is better. After the model is trained, When the test set is full of normal sentences, the higher the probability of the trained model on the test set, the better of the model. The formula is as follows:

$$PP(W) = P(w_1w_2...w_N)^{-\frac{1}{N}} = \sqrt[N]{P(w_1w_2...w_N)} \quad (9)$$

According to the formula, the greater the sentence probability, the better the model and the smaller the degree of perplexity.

### 3.7 Emotional data source and its pre-preprocessing

The emotional data are collected from the data set collected by Minlie Huang, et.al [21], including more than 40000 sentences, which are divided into six categories: others, likes, sadness, disgust, anger and happiness, with labels from 0 to 5. The data comes from NLPCC Emotion Classification Challenge and manual annotation after filtering microblog data. Because many of the data comes from the network and has many other messy symbols, it needs preprocessing and data cleaning.

Convert sentences into digital sequences, and unify the length of sentences. Because sentences have their own length, if the benchmark length is too long, there will be many zeros in the content, and if it is too short, the meaning of the sentence will be truncated. Therefore, it is generally necessary to actively select and test a better length according to the data.

The data set is divided into training set and test set according to the proportion of 1:4.

## 4 Experimental Results and Analysis

### 4.1 Experimental environment

Table 1  Experimental environment table

| environment | Version and parameters |
|---|---|
| Operating system | Windows10 |
| Compile language | 3.7.0 |
| Deep Learning Framework | TensorFlow-gpu 1.14.0 |
| Compile Tool | Pycharm 2020.3.5 |
| GPU | NVIDIA Tesla V100 16G |

### 4.2 Experimental results

When a user inputs a text, the system generates poem according to the input content. The following are some of the generated results.

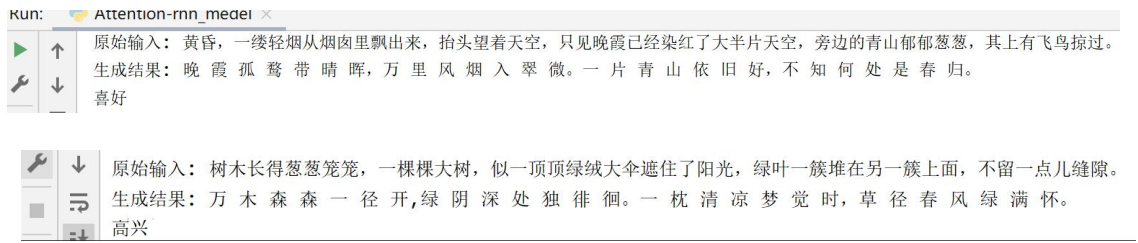

Fig.4  Result a

Fig.5  Result b

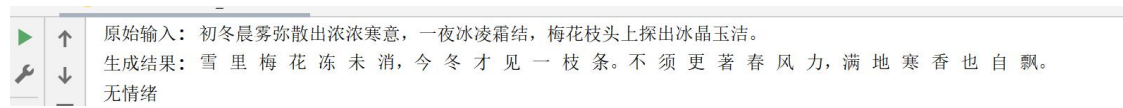

Fig.6  Result c

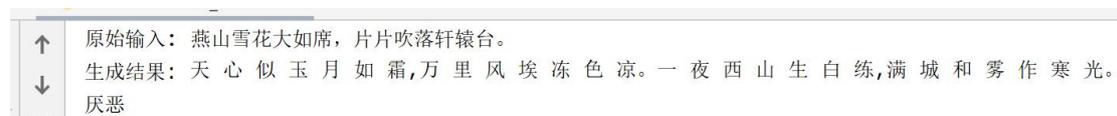

Fig.7  Result d

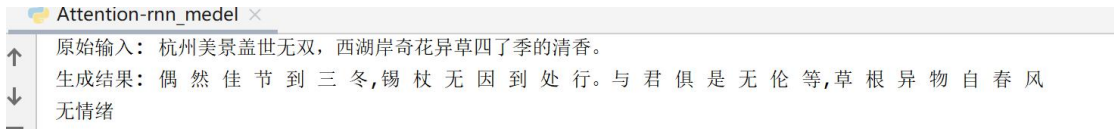

Fig.8　Result e

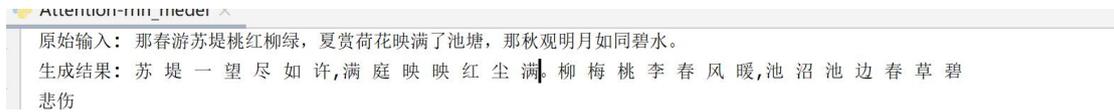

Fig.9　Result f

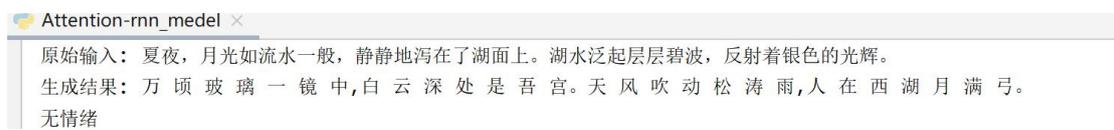

Fig.10　Result g

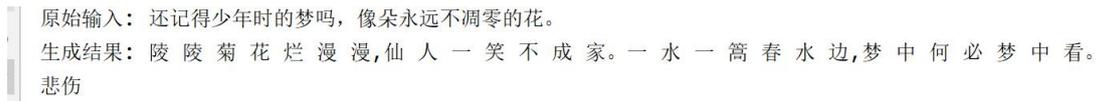

Fig.11　Result h

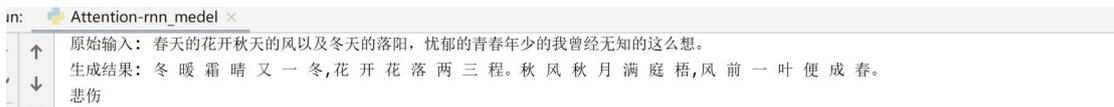

Fig.12　Result i

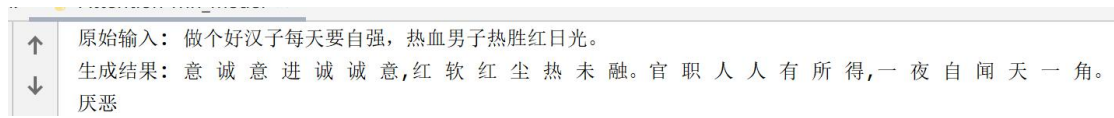

Fig.13　Result j

The generated poems are quite good.

## 5　Conclusions

This paper mainly studies the application of deep learning technology to the automatic generation of Chinese traditional poetry. The key words of each poem are compared with the existing previous text for training, so that the trained model can be generated sentence by sentence while keeping the theme unchanged. At the same time, the poetry emotion judgment function is added, which can make the poetry emotion meet the needs of users as much as possible. Because there is no poetry emotion data set that has been emotion classified, we use other article emotion classification data sets, and the effect is not satisfactory in the actual test. In addition, there are many skills in poetry, such as metaphor, comparison, etc. it is difficult to use machine generation alone, which still needs more in-depth research.